\documentclass[runningheads]{llncs}
\usepackage{times}
\usepackage{soul}
\usepackage{url}
\usepackage[hidelinks]{hyperref}
\usepackage[utf8]{inputenc}
\usepackage{graphicx}
\usepackage{todonotes}

\usepackage{hyperref}

\usepackage{subfigure}
\usepackage[nottoc]{tocbibind}
\usepackage{helvet}
\usepackage{courier} 
\usepackage{url}
\usepackage{prftree}

\usepackage{type1cm}         
\usepackage{amsmath}
\usepackage{makeidx}         
\usepackage{graphicx}        
\usepackage{multicol}        
\usepackage{tikz} 

\newcommand{\claudia}[1]{\color{black}{#1}\color{black}}

\begin{document}
\title{Consciousness and Automated Reasoning\thanks{Supported by the German Research Foundation (DFG) under the grants
SCHO~1789/1-1 and FU~263/19-1 \emph{CoRg} -- \underline{Co}gnitive \underline{R}easonin\underline{g}.}}

\author{Ulrike Barthelme{\ss}\inst{1}
 \and
Ulrich Furbach \inst{2}\and
Claudia Schon\inst{2}
}
\authorrunning{U. Barthelme{\ss} et al.}
%
\institute{  \email{ubarthelmess@gmx.de}\and University of Koblenz, Germany
\email{\{uli,schon\}@uni-koblenz.de}}
%
\maketitle              

 \raggedbottom



\begin{abstract}This paper aims at demonstrating how a first-order logic reasoning system in combination with a large knowledge base can be understood as an artificial consciousness system. For this we review some aspects from the area of philosophy of mind  and in particular Tononi's Information Integration Theory (IIT) and  Baars' Global Workspace Theory. These will be applied to the reasoning system Hyper with ConceptNet as a knowledge base within a scenario of commonsense and cognitive reasoning. Finally we demonstrate that such a system is very well able to do conscious mind wandering.
\end{abstract} 
\section{Introduction}
Consciousness and Artificial Intelligence (AI) is an  old and still ongoing debate. The question whether an AI system is able to understand in a conscious way what it is doing was very prominently raised by Searle's chinese room experiment \cite{Searle80minds}.
This is based on a symbol processing system, the Chinese room, which  is a kind of production rule system. There also is a thread of this discussion, using sub-symbolic arguments, by dealing with artificial neural networks as discussed together with various other arguments in \cite{sep-chinese-room}.

Embedding the discussion of AI systems and consciousness in a larger context, it is worth noting that  it is a topic since the mind-body problem raised by Descartes. He  postulated that mental process are properties of the mind only --- the body has to be considered separately.
The discussion since Descartes has resulted in a constant change of naturalistic (or physicalistic) and idealistic positions. 

The naturalistic position received a lot of impetus by the successes of neuroscience in connection with the ambition of artificial intelligence to model neural networks in artificial systems. 
In this position the spiritual dimension of the human being is subordinated to the physical dimension, the spiritual is derived from the physical, in short: We are only a bunch of neurons \cite{crick:95} and our ego is only an illusion, as postulated by Metzinger \cite{metzinger:ego}.

In contrast to this naturalistic view there is a long tradition of approaches which aim at understanding consciousness by taking body and mind into account. These approaches can be summarized by the notion of panpsychism, which will be discussed in Section~\ref{sec:panpshchism}.

McDermott is commenting in \cite{mcdermott:2007}, that AI researchers ``tend to shy away from questions about consciousness....The last thing most serious researchers want is to be quoted on the subject of computation and consciousness.'' 
In this paper we aim at demonstrating that an AI system, in our case  Hyper \cite{BaumgartnerFurbachPelzer2007}, an automated reasoning system for first-order logic, can be very well interpreted as a conscious system according 
 to two widely accepted theories of consciousness, namely the Information Integration Theory of Tononi \cite{tononi:04} and the 
the Global Workspace Theory (GWT) \cite{Baars:97} of Baars.
GWT, also called Baars' theater, is an approach to consciousness which is mainly motivated by the need  for  handling the huge amount of knowledge in memory.
The reasoning system Hyper resembles this model in astonishing ways, without having been specifically developed that way. 
Its development was driven only  by the need to be used in the area of deep question answering \cite{DBLP:journals/aicom/FurbachGP10,DBLP:conf/mates/FurbachS16}. For this kind of application the reasoning system has to deal with huge amounts of  general world knowledge --- the methods used to this end, make it very similar to the GWT. To push the comparison with GWT still further we examine the process of mind wandering within the Hyper system.

The first part of the paper contains a general review of issues on consciousness: The discussion around  physicality and panpsychism is depicted in Section~\ref{sec:panpshchism} and    Section~\ref{sec:info} focuses on information theoretic approaches to consciousness. 
The second part  instantiates these approaches by looking at the Hyper reasoner from the perspective of the GWT (Section~\ref{sec:reasoning}) and by introducing mind wandering with Hyper in Section~\ref{sec:mindwandering}.

\section{Physicalism versus Panpsychism}\label{sec:panpshchism}

In this section  we discuss an  approach to consciousness that allows us to attribute it  to artificial systems as well. We follow the argumentation of Patrick Spät \cite{spaet:diss}, who introduced a so called gradual panpsychism, which  is  contrasted to  physicalism. Spät argues against physicalism by stating that it   cannot really clarify what the physical is, although it claims that all phenomena of the world are based on purely physical properties: 
If physicalism  is based on the state of knowledge of contemporary physics, the definition must inevitably be incomplete, since contemporary physics does not yet provide a complete description of all phenomena occurring in the cosmos. In addition, contemporary physics may prove to be wrong. Also  reference  to a future, complete physics, which is equivalent to a ``theory of everything'', is not helpful because it   cannot be specified  further. 
According to Sp\"{a}t, a physicalist resembles Baron Münchhausen, who claims that he can pull himself out of a swamp by his own hair. For while the Baron needs an external point of reference at which he finds support and by which he can pull himself out, the physicalist needs his subjective perspective in order to have access to the world at all. Since only phenomena that can be objectively verified and formulated in the language of mathematics are to flow into the scientific description of reality, the subjective perspective is faded out. 

There are facts that go beyond the explanatory models of physicalism. In order to know what conscious experiences are and how they feel, a subjective, i.e. ``inner'' perspective of experience is required.  This is what Thomas Nagel's question aims at: What is it like to be a bat? \cite{10.2307/2183914} The physicalist can cite all kinds of physical facts about the bat, but she cannot take its perspective or experience. She cannot take it because she is not a bat and does not have the body she needs to experience the bat's  consciousness. 
However, as a result of Descartes' separation of body and mind (cogito, ergo sum), the body was isolated from the mind and regarded as an object.  This dualistic division does not take into account the fact that one experiences one's environment through and with one's body.  
The body is therefore not an ``illusion'' of the brain. Rather, the brain needs the body in order to experience its environment. Maurice Merleau-Ponty describes the conscious experience and physical ``immersion''  in the outside world using a footballer who does not perceive the ball, the boundaries of the playing field and the goal as representational or analytical --- rather, the football field is  ``present as the immanent target of his practical intention'' by actively acting in the football field \cite{merleau2}.
 When we communicate with other people, we can --- or we try to --- read from their gestures and facial expressions how they are doing.
Consciousness is entangled with the body and our experience shows that we can approach the consciousness of other beings, but this becomes more and more difficult the further we genetically distance ourselves from each other (see the bat above).   

Spät introduced the concept of ``gradual panpsychism''  by assuming a fundamental connection between mind and matter. He is arguing that  a  specificity of mind goes hand in hand with the complexity of matter or organisms. 
He assumes a very simple rudimentary form of mentality, namely mind as an ability to process information. It is based on Bateson's concept of information: A ``bit''  of information is definable as a difference which makes a difference \cite{Bateson:72}. Information is a difference that changes the state of a system, i.e. creates another difference. As soon as several differences exist in a system, a selective operation is necessary, i.e. a decision must be made.
Following Whitehead's demand ``that no arbitrary breaks may be introduced into nature'' \cite{Whitehead:67}, panpsychism is assuming that not only humans and animals, but also cells, bacteria and even electrons have at least rudimentary mental properties.
Of course there are numerous critics to this kind of understanding consciousness. John Searle  vehemently criticizes this view in \cite{Searle:2013}: "Consciousness cannot spread over the universe like a thin veneer of jam; there has to be a point where my consciousness ends and yours begins." We will further discuss this critics in the following section on Information Integration Theory.

%

\section{Information and Consciousness}
\label{sec:info}

This section discusses  two  information-based approaches to consciousness, which are defined along the lines of panpsychism from the previous section. One is the information integration theory of Tononi \cite{tononi:04} and the other, the global workspace theory of Baars \cite{Baars:97},   can be seen as 
an instance of Tononi's theory.

\subsection*{Information Integration}
Information integration theory (IIT) of Tononi \cite{tononi:04} avoids the necessity of a neurobiological correlate of consciousness. It is applicable to arbitrary networks of information processing units, they need not to be neural or biological. Tononi is proposing a thought experiment: Assume you are facing a blank screen, which is alternatively on and off and you are instructed to say ``light'' or ``dark'' according to the screen's status. Of course a simple photodiode can do exactly the same job, beep when the light is on and silence when it is off. The difference between you and the photodiode is the so called ``qualia''  --- you consciously experience ``seeing''  light or dark. This is a partially  subjective process, a first-person feeling, which we are not able to measure or compare  with that of other persons (a prominent treatment of this topic is in \cite{10.2307/2183914}). One difference between you and the photo diode is, that the diode can switch between two different states, on and off, exclusively whereas your brain enters one of an extremely large number of states when it recorgnizes the light. But it is not just the difference in the number of states, it is also important to take the degree of information integration into account. If we use a  
megapixel camera instead of a single photodiode for differentiating light from dark, this technical device would also enter one of a very large number of possible states (representing all possible images it can store). According to Tononi the  difference between you and the camera, is that the millions of pixels within the camera are not connected to each other. Your brain, however, integrates  information from various parts of the brain. (For example it is hard to imagine colours without shapes.) Tononi gives in \cite{tononi:04} a formal definition of information integration by defining a function $\Phi$, which measures the capacity of a system to integrate information. 
To get an idea of this approach, assume a network of elements which are connected, e.g. a neural network and take a subset $S$ from this system. Tononi "wants to measure the information generated when $S$ enters a particular state out of its repertoire, but only to the extent that such information can be integrated, i.e. it can result from causal interactions within the system. To do so, we partition $S$ into $A$ and its complement $B$ \ldots We then give maximum entropy to the outputs of $A$, i.e. substitute its elements with independent noise sources of constrained maximum variance. Finally, we determine the entropy of the resulting responses of $B$ \ldots" \cite{tononi:04}. Based on this computation  he  defines the \emph{effective information between $A$ and $B$}, which is a measure  of the information shared between the source $A$ and the target $B$. The above mentioned function $\Phi$ is defined with the help of the notion of effective information. The entire system is then divided into several bipartitions in order to find the ones  with the   highest  $\Phi$-value. These so-called complexes are responsible for integration of information within the system. Tononi is considering them  as the "subjects" of experience, being the locus where information can be integrated.

Based on this understanding of consciousness, one can try to test the theory by considering  several neuroanatomical or neurophysiological factors that are known to influence consciousness of humans. Tononi is doing this in great detail in \cite{tononi:04}, others are developing new Turing tests for AI systems based on this theory (e.g. \cite{Koch:2008}).

It is not surprising that IIT is criticized with similar arguments as those used against panpsychism. In the aforementioned critic of Searle \cite{Searle:2013} there is an  additional argument with respect to the notion of information. Searle is arguing that information according to Shannon is observer-dependent, it is 'only' a theory about syntax and encoding of contents, whereas consciousness is ontologically subjective and observer-independent. Koch and Tononi's reply  \cite{Koch:Tononi:2013}  is that they use "a non-Shannonian notion of information—integrated information —- which can be measured as "differences that make a difference" to a system from its intrinsic perspective, not relative to an observer. Such a novel notion of information is necessary for quantifying and characterizing consciousness as it is generated by brains and perhaps, one day, by machines." This is exactly why we think that IIT is very well suited to investigate and to experiment with aspects of consciousness in artificial systems.
We will comment later on applying  IIT to the   automated reasoning system  Hyper.

In the following  subsection we will depict another approach to consciousness, which can be seen as a special case of the information integration theory, namely the global workspace theory developed by B. Baars \cite{Baars:97}.

\subsection*{The Theater of Consciousness}
\label{sec:GWT}
One motivation for Baars global workspace theory (GWT) \cite{Baars:97} is the observation, that the human brain has a very limited working memory. We can actively  manipulate  about seven separate things at the same time in our working memory. 
This is an astonishing small number in contrast to the more than 100 billion neurons of the human brain. Another limitation is that human consciousness is limited to only one single stream of input. We can listen only to one speaker at a time, we cannot talk to a passenger during driving in heavy traffic and there are many more examples like this. At the same time there are numerous processes running in parallel but unconsciously. GWT is using the metaphor of a theater to
 model how consciousness enables us  to handle the huge amount of knowledge, memories and sensory input the brain is controlling at every moment.

GWT assumes a theater consisting of a stage, an attentional spotlight shining at the stage, actors which represent the contents, an audience and some people behind the scene. Let's look at the parts in more detail:

\emph{The stage.} The working memory consists of verbal and imagined items. Most parts of the working
 memory are in the dark, but there are a few active items, usually the short-term memory. 

\emph{The spotlight of attention.} This bright spotlight helps in guiding and navigating through the working memory. Humans can shift it at will, by imagining things or events.

\emph{The actors} The actors are the members of the working memory; they are competing against each other to gain access to the spotlight of attention.

\emph{Context behind the scene.}
Behind the scenes, the director coordinates the show and stage designers and make-up artists prepare the next scenes. 

\emph{The audience.} According to Baars, the audience represents the vast collection of specialized knowledge. It can be considered as a kind of long-term memory and consists of specialized properties, which are unconscious. Navigation through this part of the knowledge is done mostly unconsciously. This part of the theater is additionally responsible for interpretation of certain contents e.g., objects, faces or speech -- there are some unconscious automatisms running in the audience.

It is important to note, that this model of consciousness, although it uses a  theatre metaphor, is very different from a model like the Cartesion theatre, as it it discussed and refused  in \cite{Dennett1992-DENTAT}. A Cartesian theater model would assume, that there is a certain region within the brain, which is the location of consciousness, this would be a kind of humunculus. In Baars theater, however, the entire brain is the theater and hence there is no special location, it is the entire integrated structure which is conscious. This is in very nice accordance with Tononis information integration theory, described above.
And indeed Baars and colleagues developed an architecture based on GWT, which they call LIDA "a comprehensive, conceptual and computational model covering a large portion of human cognition" \cite{Baars:LIDA}. 
LIDA consists of a number of software modules, which implement a cognitive cycle which is derived from the GWT. Baars gives a detailed justification of LIDA by modelling aspects of human cognition within his model. There are numerous other cognitive architectures which are used for different purposes, an extensive overview can be found in \cite{kotseruba202040}.

In the following we will follow a very different road --- instead of designing a new architecture based on GWT, we will show that an existing automated reasoning system for first-order logic, namely the Hyper-System \cite{BaumgartnerFurbachPelzer2007} as it is used within the CoRg --- Cognitive Reasoning project \footnote[1]{http://corg.hs-harz.de} \cite{DBLP:journals/ki/SchonSS19}
can be seen as an instance of GWT. It is important to note, that the development of this system was driven by the needs to be used as a reasoner within a commonsense system together with huge amounts of world knowledge. This system was not developed as a model for consciousness, but exhibits nevertheless strong similarity with the GWT and the IIT. It can be quite understood as support of the theory used there.

\section{Automated Reasoning and GWT}
\label{sec:reasoning}
\claudia{In the CoRg project, we tackle commonsense reasoning benchmarks like the Choice of Plausible Alternatives (COPA) Challenge \cite{roemmele2011choice} with the help of automated reasoning. These problems require large amounts of background knowledge.  Similar to commonsense reasoning, the GWT assumes large amounts of background knowledge. Therefore, the methods developed in the CoRg project are suitable for modelling of the GWT by means of automated reasoning.
In this section, we not only briefly depict an automated reasoning system but also comment on particular problems that arise if it is applied in the context with large knowledge bases. }In Subsection~\ref{sec:gwt-glasses} we interpret the system along the lines of Section~\ref{sec:GWT} as a Baars' theater and hence a model the GWT.

\subsection{Reasoning with Large Knowledge Bases}
\begin{figure}[t]
  \begin{minipage}{4 cm}

\begin{tikzpicture} 
  \node [name = n1,align=left,anchor=south]{$dog(a)$} [sibling distance = 2.cm, node distance = 3cm,level distance=0.75cm]    	
    child {node {$bone(b)$} [sibling distance = 2.cm, node distance = 3cm,level distance=0.75cm]    	
    child {node {$chew(c)$} [sibling distance = 2.cm, node distance = 3cm,level distance=0.75cm]    
    child {node {$on(c,b)$} [sibling distance = 2.cm, node distance = 3cm,level distance=0.75cm]    	
    child {node {$\mathit{agent(c,a)}$} [sibling distance = 2.cm, node distance = 3cm,level distance=0.75cm]    
    child {node {$manducate(c)$} [sibling distance = 2.cm, node distance = 3cm,level distance=0.75cm]    	
    child {node {$eat(c)$} [sibling distance = 2.cm, node distance = 3cm,level distance=0.75cm]    	
    child {node {$animal(a)$} [sibling distance = 2.cm, node distance = 3cm,level distance=1cm]    	
        child {node {$herbivore(a)$}  [sibling distance = 2.cm, node distance = 3cm,level distance=1cm]    	
      child {node {$\begin{array}{c} plant(b) \\ \bot \end{array}$}} 
    }
    child {node {$carnivore(a)$} [sibling distance = 2.cm, node distance = 3cm,level distance=0.75cm]    	
    child {node {$dog\_treat(b)$} [sibling distance = 2.cm, node distance = 3cm,level distance=0.75cm]    	
    child {node {$\mathit{dog\_food(b)}$} [sibling distance = 2.cm, node distance = 3cm,level distance=0.75cm]    	} [sibling distance = 2.cm, node distance = 3cm,level distance=1cm]    	
    }
    } 
    }}}}}}}
    ; 
\end{tikzpicture}

 \end{minipage}
 \hskip .18 cm
 \begin{minipage}{8 cm}
\begin{align}
&dog(a) \leftarrow \\
&bone(b)  \leftarrow \\
&chew(c)  \leftarrow \\
&on(c,b)  \leftarrow \\
&\mathit{agent(}c,a)  \leftarrow \\
&manducate(X) \leftarrow  chew(X)\\
&eat(X) \leftarrow  manducate(X)\\
&animal(X) \leftarrow  dog(X)\\
&herbivore(X) \lor carnivore(X) \leftarrow  animal(X)\\
&plant(Z) \leftarrow  herbivore(X) \land manducate(Y) \land \nonumber \\
&\phantom{plant(Z) \leftarrow\ }\mathit{agent}(Y,X)  \land on(Y,Z)\\
&\bot\leftarrow  plant(X) \land bone(X)\\
&dog\_treat(X)\leftarrow  bone(X)\\
&\mathit{dog\_food}(X)\leftarrow  dog\_treat(X) 
\end{align}

\end{minipage}
\caption[bla]{\small Clauses on the right, hyper tableau for the clauses on the left. 
The left branch of the tableau is closed, the right branch is open. The literals in the open branch constitute a partial interpretation of the clause set.
    }
  \label{F:proof2}
  \end{figure}
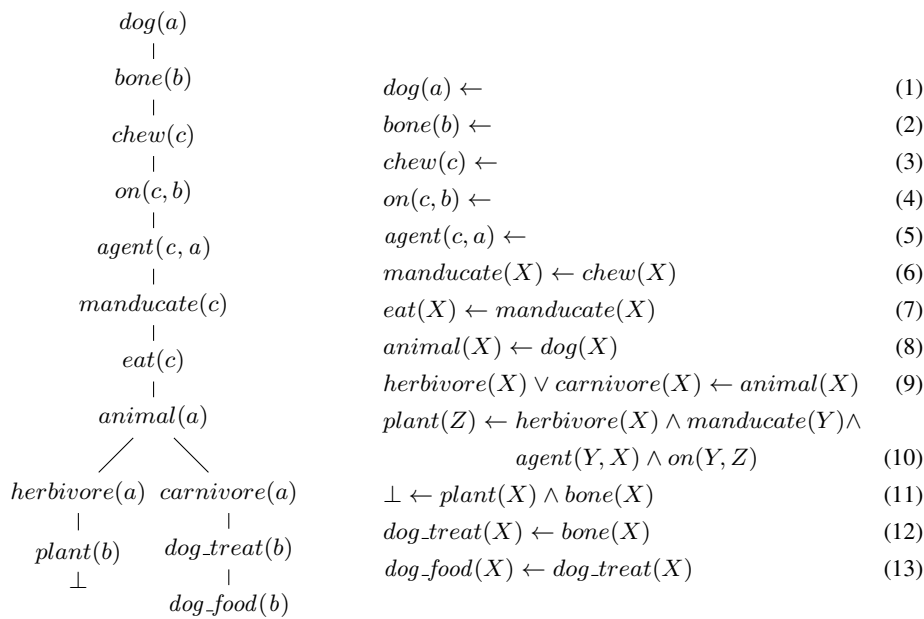

Hyper is an automated theorem prover for first-order logic \cite{BaumgartnerFurbachPelzer2007}. First-order theorem proving is aiming at the following task: Given a set of formulae $F$  and a conjecture (sometimes called query) $Q$ the question is, whether or not $Q$ is a logical consequence of $F$, written as $F \models Q$. For first-order logic formulae this is an undecidable problem, but it is semi-decidable, meaning that if the logical consequence holds, the prover will stop after finite time stating that it is a consequence. Hyper, like most of the high performance provers today is a refutational prover, which means that the question whether  $F \models Q$ holds, is transformed into the equivalent question asking if $F \cup \lnot Q$ is unsatisfiable. Before trying to prove the unsatisfiability of $F \cup \lnot Q$, Hyper transforms $F \cup \lnot Q$
into a normal form, a so-called set of clauses. The Hyper prover is based on a tableau calculus. The advantage of this is, that Hyper is manipulating one single proof-object, the tableau,  in order to demonstrate the unsatifiablity of the problem at hand. 
Figure~\ref{F:proof2} shows an example of a clause set belonging to a problem from the area of commonsense reasoning together with its Hyper tableau. The tableau on the left part of Figure~\ref{F:proof2} is essentially a tree that was developed branch by branch using inference steps. An inference step selects a branch and then tries to extend this branch by using a clause from the clause set (right side of Figure~\ref{F:proof2}) together with an inference rule specified by the calculus. The technical aspects of the calculus on how to extend the tree in detail are not important here.  We want to point out, however, that
at any stage of the construction of a tree, a branch represents a (partial) interpretation of the given formulae. E.g., the right-most branch in our example corresponds to the (partial) interpretation $\{dog(a), bone(b),chew(c),on(c,b), \mathit{agent(c,a}), manducate(c), eat(c), animal(a),$\\
$carnivore(a), dog\_treat(b), \mathit{dog\_food}(b)\}$, 
The left branch of the tableau is closed, since $bone(b)$ and $plant(b)$ (derived using clause (10) with $X=a$, $Y=c$ and $Z=b$) are contradictory according to clause (11). 
A proof is found if there is a tableau that contains only closed branches. If no proof can be found, like in the example in Figure~\ref{F:proof2}, the open branches list literals that can be derived from the set of clauses.

Hyper has been used in many different application areas, reaching from commercial knowledge based systems to intelligent book development \cite{DBLP:journals/jar/BaumgartnerFGS04}. 
Recently Hyper was used as the main reasoning machinery in natural language query answering \cite{DBLP:journals/aicom/FurbachGP10} and for cognitive reasoning, in particular answering commonsense questions \cite{DBLP:conf/mates/FurbachS16}.

In all of these applications Hyper very rarely managed to find a proof within the given constraints --- in most cases there was a timeout and Hyper's result was a branch representing a partial interpretation of the formulae at hand. 

Next, we will illustrate how to use Hyper to draw inferences from a statement. Since we aim at modeling the GWT, we assume the statement to be given in natural language. 
In order to draw inferences from a natural language statement with the help of an automated reasoning system, the problem has to be translated from natural language to a formal language --- in our case this is first-order logic. This translation is done in a fully automated system called KnEWS \cite{knews}, which is based on  the Boxer-System  \cite{CurranClarkBos2007ACL}.
As a running example, we consider the following sentence, which is a part of one of the problems in the COPA challenge:
\begin{quote}
The dog chewed on a bone.
\end{quote}
The first-order logic translation we get using KnEWS is:
\begin{align}
 \exists A (& dog(A) \land \exists B,C
                        (r1on(C,B)
                        \land bone(B)
                                             \land r1agent(C,A) 
                        \land chew(C)
                         )).\label{equ:chew}
\end{align}

It is obvious that for reasoning  with this formula knowledge about the world  is necessary; e.g., about food intake of dogs or about the composition of meat and bones. This knowledge of course cannot be added by hand, it is necessary to use a knowledge base, where many possible facts and relations about the world are available. If this knowledge is not restricted to a single domain, if it is general enough to be used for different areas it gets very large and hence difficult to handle. In the CoRg-project 
\cite{DBLP:journals/ki/SchonSS19}  
among other sources ConceptNet \cite{DBLP:conf/aaai/SpeerCH17} is used as background knowledge. ConceptNet is a semantic net structure with 1.6 million edges connecting more than 300,000 nodes.
Knowledge in ConceptNet is stored in the form of triples such as \emph{(dog, hasA, fur)}. To allow the first-order logic reasoner Hyper  to use ConceptNet as background knowledge, we have translated most of the English part of ConceptNet to first-order logic. The above triple has been translated into the following formula:
\begin{equation}
\forall X (dog(X) \rightarrow \exists Y (hasA(X,Y) \land \mathit{fur}(Y)))\label{equ:fur}
\end{equation}
%
The resulting knowledge base consists of 2,927,402 axioms and is therefore far too large to be completely processed by reasoners.
Hence, it is necessary to select parts of this huge knowledge base which might be relevant for the task at hand. Note, that this situation is very different from a classical automated reasoning problem, 
where all the necessary formulae to find a proof are given and can be used all together  by the reasoning system, without the necessity to guess parts of it to be loaded into the system. 

\begin{figure*}[t]
\includegraphics[scale=0.42]{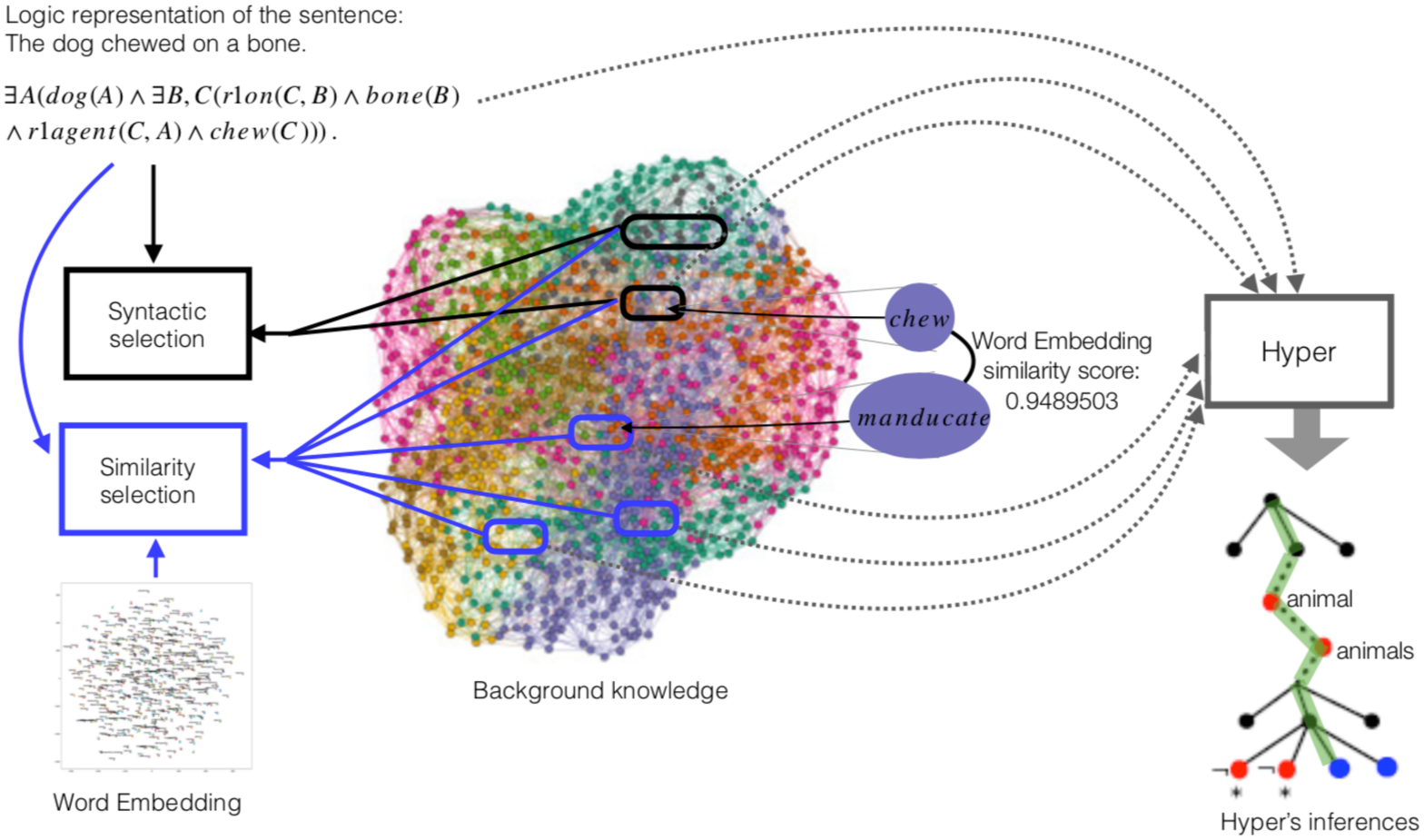}
\caption{On the left: Syntactic selection uses symbols from the formula to select parts of the background knowlege, depicted with black arrows and regions. Similarity selection takes the meaning of symbol names  into account by additionally selecting formulae containing symbols which are similar according to a word embedding (depicted by blue arrows and regions). On the right: A snapshot during a Hyper run. The (green) path of the tree, Hyper is working on, corresponds to the working memory. The knowledge base represents the long-time memory.\protect\footnotemark}
\label{fig:selection}
\vspace{-0.35cm}
\end{figure*}

The left part of Figure \ref{fig:selection} illustrates the situation in automated reasoning with large knowledge bases. The logical representation (Formula~(\ref{equ:chew})) of the natural language sentence is depicted on the very left together with the knowledge base, ConceptNet, in the middle of the figure. The task is to select those parts from the knowledge base which might be helpful for reasoning about the logical representation. To this end there are two selection methods sketched: The first one uses syntactic criteria exclusively for the selection \cite{Hoder:2011uq}. Depending on the symbols occurring within the logical representation those parts of the knowledge base are selected, which contain one of these symbols (additionally this selection takes the number of occurrences of a symbol into account in order to prevent that very frequent symbols like \emph{isA} lead to the selection of the whole knowledge base). The second method uses additionally  semantic criteria for the selection. The semantics of a symbol is given by a word embedding, which is used to find semantically similar symbols for the selection process. As a result not only formulae containing the symbols \emph{dog}, \emph{chew} and \emph{bone} are selected, but also those containing similar symbols like for example \emph{manducate} and \emph{remasticate}.
This method is described and evaluated in detail in \cite{DBLP:conf/cade/FurbachKS19}.

\claudia{Figure~\ref{F:proof2} depicts an extract of the selected background knowledge for Formula~(\ref{equ:chew}) on the right hand side:  Clauses (6) - (13) correspond to selected background knowledge for the symbols in Formula~(\ref{equ:chew}) and exemplary for the symbol \emph{manducate} which is similar to \emph{chew}. Clauses (1) - (5) correspond to the clausification of Formula~(\ref{equ:chew}). 

Even if much background knowledge is added, this background knowledge will never be able to represent the complete human background knowledge and will always remain incomplete. Therefore automatic theorem provers can only rarely prove inferences in natural language. For example, it would hardly be possible with an automatic theorem prover to prove that the statement \emph{The dog chewed on a bone} implies the statement \emph{The dog is content}. This is because the second statement is not  a logical consequence of the first one. It is rather the case, that \emph{The dog is content} is more likely to be a a consequence than the statement \emph{The dog is injured}.
This kind of reasoning is also called cognitive reasoning.


 Therefore, we do not aim at the construction of proofs but rather to analyse the inferences performed by Hyper  after a certain amount of reasoning (see the green path in the right part of Figure~\ref{fig:selection}). }
 
One of the main problems in the above depicted task is the selection of appropriate parts of knowledge. This is very much related to the approach of information integration  according to  Tononi: Given a huge network, the knowledge base, and a problem, we want to integrate all the necessary parts of the knowledge to solve the problem. The network is formed by the connections within the used knowledge bases, but also by the similarities defined by the word embeddings. Hence the degree of consciousness of the entire system from Figure~\ref{fig:selection} could be determined (at least in theory) by Tononis approach.
\color{black}

\subsection{Looking through the GWT-Glasses}
\label{sec:gwt-glasses}

In the previous subsection we briefly explained how the theorem prover Hyper is adapted and used within the area of cognitive reasoning. We now show that Hyper in combination with large knowledge bases can be interpreted as an architecture that implements the GWT as introduced in Section~\ref{sec:GWT}. 


\emph{The stage.} The working memory is the branch of the tree which currently is expanded. In the right part of Figure~\ref{fig:selection} this is the green path of the tree --- it contains the context in which the next reasoning step will be performed.

\emph{The spotlight of attention.} This bright spotlight selects and highlights those parts of the (green) branch together with the formulae from the problem or the selected parts of the  knowledge base which are used for the next  reasoning step.

\emph{The actors.} The actors correspond to the application of inference rules on the set of clauses currently processed by the theorem prover. The result of the actors' actions correspond to new formulae derived by an inference step. The spotlight of attention is deciding which actor, i.e. inference rule together with the necessary formulae, from the stage are to be active next.

\emph{Context behind the scene.}
Behind the scenes,  the reasoner and its control act as a director.

\emph{Audience.} According to Baars, the audience represents the vast collection of specialized knowledge. It can be considered as a kind of long-term memory, namely the   knowledge base. We depicted several selection mechanisms for finding the appropriate parts of the knowledge. These can be seen as the unconscious
interpretation skills which Baars located in the audience. In our case we only deal with declarative knowledge, if we had procedural knowledge as well, this would be part of the audience as well.

Altogether we have a complete Baars' theater of consciousness consisting of the reasoner Hyper, together with its control and its background knowledge --- we have a system which can be interpreted as a conscious system according to the ideas presented above. However it should be noted that "Metaphors must be used with care. They are always partly wrong, and we should keep an eye out for occasions when they break down. Yet they often provide the best starting point we can find." \cite{Baars:97}

 In the following section this will be deepened in the discussion of a conscious reasoning system in the sense of IIT by discussing mind wandering.

\footnotetext{Picture of network: \cite{DBLP:journals/asc/GibsonV16}, CC BY 4.0 (\url{https://creativecommons.org/licenses/by/4.0/}).
Visualization of Word Embedding: \emph{Euskara: Hitz batzuen errepresentazioa} by Aelu013, CC BY-SA 4.0 (\url{https://creativecommons.org/licenses/by-sa/4.0/deed.en}) (word removed).}

\section{Mind Wandering}

\label{sec:mindwandering}

Mind wandering is a process in which people do not stick to a single topic with their thoughts, but move in chains of thoughts from one topic to the next. In doing this  the border between conscious and unconscious processing is continuously crossed, and in both directions. Hence studying mind wandering certainly contributes to a better understanding of consciousness. Mind wandering often occurs in less demanding activities.  A study  \cite{killingsworth:gilbert:2010} shows that up to 40\% of the time a human mind is wandering around.  

Mind wandering also has interesting positive effects, which is  investigated in \cite{article}
where it was shown that mind wandering can be helpful in finding creative solutions to a problem. In this section we show, that a control system of Hyper is very well able to invoke mind wandering for Hyper. We will first give a rough overview of the system and then go into details about the individual steps.

\subsubsection{Overview of the System}
The mind wandering process is started from an initial formula, such as Formula~(\ref{equ:chew}). In the first step, the system performs a semantic selection as described in the previous section to select suitable background knowledge for this formula. The formula together with the selected background knowledge is transferred to Hyper, which performs inferences and returns a (possibly partial) interpretation. This (possibly partial) interpretation corresponds to the green path of the tree in the right part of Figure~\ref{fig:selection} and contains everything Hyper was able to derive within a given time limit. Since the selected background knowledge is very broad, the interpretation also contains very broad information. To find a focus, the symbols occurring in the interpretation are clustered according to their word meaning and a cluster is selected as focus. 
This step simulates the \emph{spotlight of attention} of the GWT, which focuses on a certain area of the stage. For the symbols in the focus, new background knowledge is selected and passed to Hyper again together with the focus. Hyper again performs inferences. This process is repeated until Hyper's inferences no longer provide new information compared to the previous round. 
A detailed description of the individual steps of the system follows.

\subsubsection{The Audience -- Background Knowledge and Selection}

 The  selection from the knowledge base  starts with a set of symbols called the current \emph{context}, which consists of the symbols from a starting formula like for example Formula~(\ref{equ:chew}) and similar symbols.

Selecting all formulae from the knowledge base in which one of the context symbols occurs results in a large set of formulae. Using all these formulae would be too unfocused w.r.t. the considered formula, so a filtering step removes all formulae in which other predicate symbols occurring in the formula are not within a certain range of similarity to the symbols in the context. To measure similarity, cosine similarity in a word embedding is used. 
The interval in which the similarity must fall for a formula to be selected is passed to the system by two parameters. With the help of these parameters it is possible to control how far the background knowledge is allowed to move away from the context symbols.  With a suitable interval it is possible to select  a formula like (\ref{equ:fur}) while preventing to select a formula like
\begin{equation}
\forall x (poodle(x) \rightarrow \exists y (\mathit{relatedTo(x,y)} \land dog(y)))\label{equ:poodle}
\end{equation}
Currently, the system can use either the ConceptNet Numberbatch \cite{DBLP:conf/aaai/SpeerCH17} word embedding or a word embedding  learnt on personal stories from blog entries \cite{roemmele2011choice}.

\subsubsection{Actors and Context behind the Scene --- Reasoning}
The selected set of formulae together with  Formula~(\ref{equ:chew}) is passed to the Hyper reasoner. Hyper is started with a timeout of 30 seconds and calculates during this time a possibly partial interpretation for the input formulae. This interpretation represents knowledge that can be inferred from Hyper's input. In the next step, the system analyses Hyper's output. Since the input formulae are still very broad despite the filter methods mentioned above, the (partial) interpretation also contains a very broad knowledge inferred from Hyper's input. 
First the system extracts all predicate symbols from Hyper's output and removes from this set all symbols from the current context to prevent the mind wandering process from getting stuck. Hyper's model produced for the running example contains 122 predicate symbols which are thematically widely spread: from \emph{ears, skin, flesh, wolf, calcium, animal, collar} and \emph{vertebrate} to \emph{woof} and \emph{barking}, many terms are represented. 

\subsubsection{Spotlight of Attention --- Finding a Focus}
To determine a focus in the multitude of these terms, the system performs a clustering on these terms using KMeans and the cosine similarity of a word embedding as similarity measure. Currently, the number of clusters created corresponds to the number of predicate symbols in the (partial) interpretation divided by 4. In future work, different values for the number of clusters will be considered.
Next, the system orders the resulting clusters by their cosine similarity to the predicate symbols in the current context and chooses one of the clusters as the focus. For the experiments, the cluster in the middle of the sorted sequence is chosen as the focus in order to allow the mind wandering process to move away from the current context. In the running example, this led to selecting the cluster consisting of the symbols \emph{animal} and \emph{animals} as the new focus. Other choices for the focus cluster are possible and can be selected with the help of a parameter.
Next, the system creates a simple formula from the symbols in the focus cluster and selects suitable background knowledge as described above. In this selection, the symbols from the focus cluster together with similar symbols are used as new context symbols.
The described process 
is repeated a desired number of times or until the process does not deliver any new symbols.

\subsubsection{Experimental Results}Starting from the symbols in the initial formula, the symbols in the selected focus clusters represent the result of the mind wandering process.
Starting from Formula~(\ref{equ:chew}) containing the symbols \emph{dog}, \emph{chew} and \emph{bone} the described system provides for example the following sequence of sets of focus symbols:

\noindent $\{$\emph{dog, chew, bone}$\}$ $\rightarrow$$\{$\emph{animal, animals}$\}$ $\rightarrow$ 
$\{$\emph{gardening}$\}$ $\rightarrow$ 
$\{$\emph{garden, horticulture, farming}$\}$ $\rightarrow$ 
$\{$\emph{mowing, lawn, yard}$\}$ $\rightarrow$ 
$\{$\emph{outside, front, outdoor}$\}$ $\rightarrow$ 
$\{$\emph{weather}$\}$ $\rightarrow$ 
$\{$\emph{thunder, lightning}$\}$ $\rightarrow$ 
$\{$\emph{cloud, sky, clouds}$\}$ $\rightarrow$ 
$\{$\emph{water}$\}$
\\
This corresponds to a mind wandering chain which focuses on animals  leads  to gardening and finally addresses weather aspects which leads to water.

It should be noted that the system has many parameters to control this mind wandering process. For the experiments, different parameter combinations were automatically tried out and the sequences of focus symbols generated in this way were manually inspected. 
Different parameter values led to a different chain which finally ends at fashion:

\noindent $\{$\emph{dog, chew, bone}$\}$ $\rightarrow$ $\{$\emph{furry, tail, fur}$\}$ $\rightarrow$ 
$\{$\emph{coats, wool, coat}$\}$ $\rightarrow$ 
$\{$\emph{fur}$\}$ $\rightarrow$ 
$\{$\emph{animal, pet, hair, coat, pelt, wool}$\}$ $\rightarrow$ 
$\{$\emph{sleeves, robe, braided, leather, fur, garment, buttoned, styled, pockets, strand, woven, cloth, wearable}$\}$ $\rightarrow$ 
$\{$\emph{coats, covering, pattern, textile, fastened, worn, wool, coat, material, pelt}$\}$ $\rightarrow$ 
$\{$\emph{wearing, robe, suit, shirt}$\}$

The presented experiments are only a first feasibility study. For future work the application of mindwandering in the commonsense reasoning area is planned.

\section{Conclusion and Future Work}

In this paper we tried to connect work from research on consciousness with work on formal reasoning. We depicted an implementation of a mind wandering process within a logical reasoning system, which can be interpreted as the action of a consciously reasoning system. Further work has to be done for finding a way to determine what knowledge  is interesting enough to be kept within the focus of the system and how the knowledge base should be modified according to the results of mind wandering. Currently, only one path of the proof tree is considered for the mind wandering process. In future work we plan to extend the approach to consider multiple open branches for mind wandering.

Furthermore, we plan the application of mindwandering in the commonsense reasoning area, where we will consider commonsense reasoning benchmarks like the Choice of Plausible Alternatives Challenge \cite{roemmele2011choice}. Figure~\ref{fig:copaproblem} shows an example from the COPA Challenge. 
\begin{figure}[h]
\emph{
65: The family took their dog to the veterinarian. What was the CAUSE of this?}
\begin{enumerate}
\itemsep-2pt
\item \emph{The dog chewed on a bone.}
\item \emph{The dog injured his paw.}
\end{enumerate}
\vspace{-3mm}
\caption{Example problem 65 from the COPA challenge.}
\label{fig:copaproblem}
\vspace{-0.35cm}
\end{figure}

A first idea for the solution of these benchmarks would be to start a mindwandering process for both answer candidates and to choose the answer where the result of the mindwandering process is closer to the sentence \emph{The family took their dog to the veterinarian}.


\bibliographystyle{splncs04}
\bibliography{wissen}

\end{document}